# Stock Market Forecasting Based on Text Mining Technology: A Support Vector Machine Method

Yancong Xie[1, 2], Hongxun Jiang[2*]

[1] Information Systems School, Queensland University of Technology, Brisbane, Australia & School of Information, Renmin University of China, Beijing, China.
[2] School of Information, Renmin University of China, Beijing, China.

* Corresponding author. Tel.: +86 10 82500904; email: jianghx@ruc.edu.cn

**Abstract:** News items have significant impact on stock markets but the ways is obscure. Many previous works have aimed at finding accurate stock market forecasting models. In this paper, we use text mining and sentiment analysis on Chinese online financial news, to predict Chinese stock tendency and stock prices based on support vector machine (SVM). Firstly, we collect 2,302,692 news items, which date from 1/1/2008 to 1/1/2015. Secondly, based on this dataset, a specific domain stop-word dictionary and a precise sentiment dictionary are formed. Thirdly, we propose a forecasting model using SVM. On the algorithm of SVM implementation, we also propose two parameter optimization algorithms to search for best initial parameter setting. Result shows that parameter G has the main effect, while parameter C's effect is not obvious. Furthermore, support vector regression (SVR) models for different Chinese stocks are similar whereas in support vector classification (SVC) models best parameters are quite differential. Series of contrast experiments show that: a) News has significant influence on stock market; b) Expansion input vector for additional situations when that day has no news data is better than normal input in SVR, yet is worse in SVC; c) SVR shows a fantastic degree of fitting in predicting stock fluctuation while such result has some time lag; d) News effect time lag for stock market is less than two days; e) In SVC, historic stock data has a most efficient time lag which is about 10 days, whereas in SVR this effect is not obvious. In addition, based on the special structure of the input vector, we also design a method to calculate the financial source impact factor. Result suggests that the news quality and audience number both have significant effect on the source impact factor. Besides, for Chinese investors, traditional media has more influence than digital media.

**Key words:** Impact factor, parameter optimization, support vector machine, stock market, text mining.

## 1. Introduction

Nowadays a heated topic is forecasting financial market. It is not exaggerated to say the foundation of a country is its economic market. And at the heart of any market economy, lies the financial markets with their supply and demand equilibriums. Within these fields, stock market attracts most eyes because of its volatility and importance. However, to predict stock market is a difficult task. Efficient market hypothesis states that it seems impossible to predict stock prices precisely and that stocks behave in random walk manner, influenced by tons of macro-economic factors such as political events, firms' policies, general economic conditions, commodity price index, bank rate, bank exchange rate, investors' expectations, institutional investors' choices, movements of other stock markets, psychology of investors, etc.

Predicting stock market began heated in 1960s using Moving Average (MA), exponential smoothing, and ARIMA [1], [2]. After that, more and more researchers are turning to nonlinear methods, based on various types or combinations of machine learning algorithms [3]. Besides, data sources also experiences a big change. Researchers are no more satisfied with historic stock data. They try to make use of different sources' data and make assumptions based on that [4].

Among these additional sources, text sources are starting to become heated because of its maneuverability [5]-[7]. Actually, text sources are also divided into two parts. One is focused on the direct feedback of customer and market reception for a new product to estimate the popularity and price of a product, even a stock. It is mainly based on identifying positive and negative words and processing text with the purpose of classifying its emotional stance as positive or negative, widely used in predict turning point or tendency of a specific stock [8]-[10]. The other one tries to deal with passive influence from external text sources. Such sources are usually from mass-media, like main websites, newspapers and magazines [9], [11]. These texts tell objective information and investors react to this information differently in their investment decision-making process.

Chinese stock market is special around the world. Unlike western countries' stock market, most stock investors are ordinary people, not investment firms and corporations. Besides, the majority of Chinese investors do not have enough financial or economic knowledge. Most of them make money by short-term speculative trading and can be easily manipulated by external information. Thus, news should have significant effect on Chinese stock market. However, till now there are not many researches trying to deal with this phenomenon. Facing so many difficulties, this research will focus on support vector algorithm using text mining method to deal with online news and predict Chinese stock market. We hope to demonstrate a new approach to describe stock market movement and uncover unique mechanism of Chinese stock investor behavior as well.

## 2. Research Method

Like many other similar researches, this research is mainly divided into 3 stages: data collection; data pre-processing; machine learning and forcasting, as shown in Fig. 1.

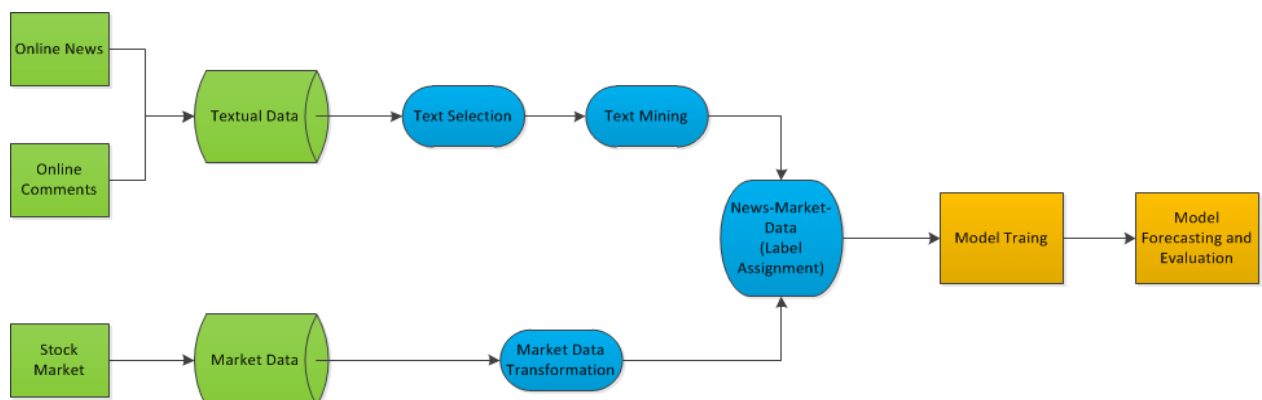

Fig. 1. General research process.

### 2.1. Data Preparation

This research merges textual data and market data together to form input vector. When collecting textual data, it is essential to find out all possible news for any stocks. There are two benefits doing so: 1) text mining dictinaries can be more precise; 2) more latent influence from text data can be captured. By contrast, stock type does not need to be overall. However, the type of testing stocks should be representative. Thus, selecting rules of the two kind data are quite different.

### 2.1.1. News source selection

When deciding news sources, the basic importance and source' field are both considered to guarantee text data's effectiveness and universality. Since main Chinese investors make their decisions relying on news in Chinese, the news language we choose here is Chinese. News related to stock market is divided into 8 parts: financial capital news; civil economic news; industrial economic news; company news; international economic news; emerging market news; consumption news; financial news. This research selects 20 sources covering the main Chinese online financial news fields.

Python-Scrapy and Mongodb are used in this research as crawling framework. 2302692 pieces of news documents dating from 1/1/2008 to 1/1/2015 are obtained.

### 2.1.2. Stock selection

This research selects 20 Chinese stocks as testing data. When selecting, trade volume is the most important factor to be considered. Stocks with high trading volume are supposed to have high mobility, which should be the main concerns of investors. Besides, having a good understanding of such kind of stocks would be a great help for a government to regulate and control its economic market. Apart from trading volume, factors like stock sectors, publishing date and stock price are also taken into consideration.

## 2.2. Data Pre-processing

The input vector includes two parts' information: online news and stock data, representing influence from the two aspects. One input vector is designed to contain affecting factors in a single day. The input vector, thus, includes $X$ text nodes and $Y$ stock data nodes. $Y$ is the conjunct time lag of stock data. $X$ is the number of news sources, representing the news influence from one specific source.

In this research, choosing text sources is mainly based on the amount of each source's news and the $X$ is set to be 20, representing the most important sources in Chinese market. How to find out the most efficient $Y$ will be discussed in the next chapter. The overall process is shown in Fig. 2.

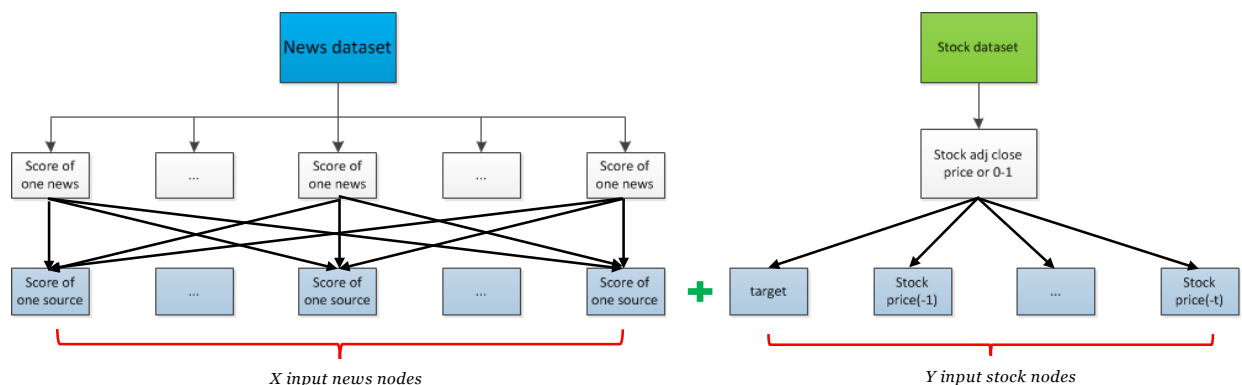

Fig. 2. Data pre-processing.

### 2.2.1. Text mining

Like other related researches, this part is divided into 4 stages, as shown in Fig. 3. Before begin to analyse stock market result, text mining work is done at first to get Stop-words Dictionary and [-5,+5] Domain-Specific Sentiment Dictionary. Later, the two dictionaries are cited in specific text mining process.

In specific, this research follows the main stream of text mining process [4], [8], [12]. The procedure here is quite elementary. First, text document is cut into a bag of words (BOW) citing exisiting dictionaries. Second, a frequency analysis is conducted to select 500 most significant words (leave out those already in Stop-words Dictionary and Domain-Specific Sentiment Dictionary). Then three experts score the result words from -5 to +5 (where –5 represents the most negative effect, +5 represents the most positive effect and 0 represents no significant meaning (stop word)). If two or more persons think one word should be a stop word, this word will be thrown into Stop-words Dictionary. Otherwise, this word will be added into Sentiment Dictionary whose sentiment score will be average of three persons. Then a new text cut turn begins. The iterative process will carry on until 90% of the selected words have been scored.

After that, citing the two updated dictionaries, structured input data is formed for each firm. The text mining results thus are prepared for the SVM experiments.

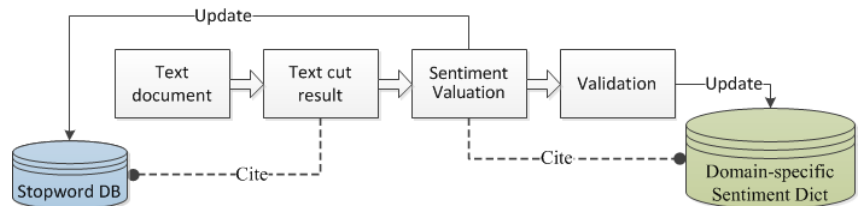

Fig. 3. Process of text mining.

Table 1. Experiment Details

|  | *Time Lag* | *Input Data Expansion* | *Traverse Algorithm* | *Approximate Algorithm* |
|---|---|---|---|---|
| **Purpose** | Seek best time lag | Investigate the influence of expansion input dataset | Parameters optimization | Parameters optimization |
| **News input** | Variance | Expansion | Standard | Standard |
| **Stock input** | Different length | Standard | Standard | Standard |
| **SVM parameters (C & G)** | Default | Default | Search for global optimum | Use local optimums to approximate global optimum |
| **Firms** | Bank of China | All | All | All |
| **Methods** | SVR & SVC | SVR & SVC | SVR & SVC | SVR & SVC |

### 2.2.2. Stock price

Stock market price falls into four categories: Open price, High price, Low price and Close price. Most related researches choose to predict Close price. However, the Close price fluctuates a lot because of multiple factors like exchange rate. Thus, this research chooses to use *Adj* Close price, which means divided-weight price, as input variable.

### 2.2.3. Support vector machine

In this research, LIBSVM is used as the machine-learning environment. C-SVC will be used to predict the tendency of stocks and polynomial kernel will be used, which is:

$$K(u,v) = (r \cdot u' \cdot v + coef0)^{\text{degree}}$$

Whereas epsilon-SVR will be used to predict the specific prices of stocks and use kernel is sigmoid kernel, shown below:

$$K(u,v) = tanh(r \cdot u' \cdot v + coef0)$$

In these two kernels, *degree* is default to be 3 and *coef0* is default to be 0.

In addition, since the numbers of news input and market input are differential, there is a need to scaling input vectors to make all numbers in the same level. Scaling function is shown as follows.

$$z = \frac{X - \overline{X}}{\sigma}$$

where $\overline{X}$ is the average of *X* and $\sigma$ is standard deviation of *X*. After scaling process, the whole input data will belong to standard normal distribution. Such kind of data will perform better in the SVM algorithm.

## 3. Experiments and Results

In this chapter, one stock will be tested to find out a common efficient parameter set. Then all stocks will be tested together to valid the model in series of contrast experiments. The chosen test stock is *Bank of China* whose company code is *601988*. A table below makes comparisons of experiments in this research.

### 3.1.1. Time lag

In this experiment, time lag is set to step from 1 to 20. There are two types of input vector. One contains news data, whereas the other includes only stock data. Support vector classification (SVC) is used to predict stock tendency and support vector regression (SVR) is applied to predict stock price. Results are shown in Fig. 4 and Fig. 5.

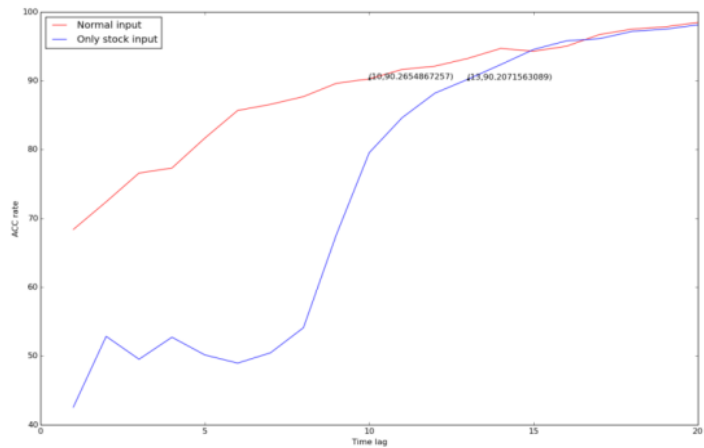

Fig. 4. SVC result for different time lag.

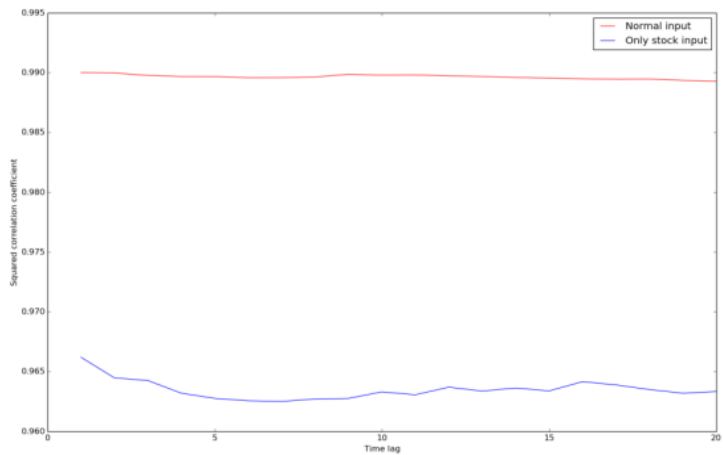

Fig. 5. SVR result for different time lag.

As can be seen from the two figures: 1) news is highly relevant with the stock market; 2) Chinese stock news has a short-term effect. In other words, stock investors can be affected by news in a special environment and in short time; 3) news has 2 days effect on stock, even shorter. Based on the result, this research choosing 10 as the most efficient time lag to predict stock tendency, deem 90% as the threshold for a good prediction. In the later experiments, we use 10 as the standard time lag without explanation.

### 3.1.2. Input expansion

It is universal to all that more overall train data set is, more robust the train model will be. Thus, we expand the train dataset by setting all news node' values to be 0 when that day has no responding news. The two kinds of input's performance are shown in Fig. 6 and Fig. 7.

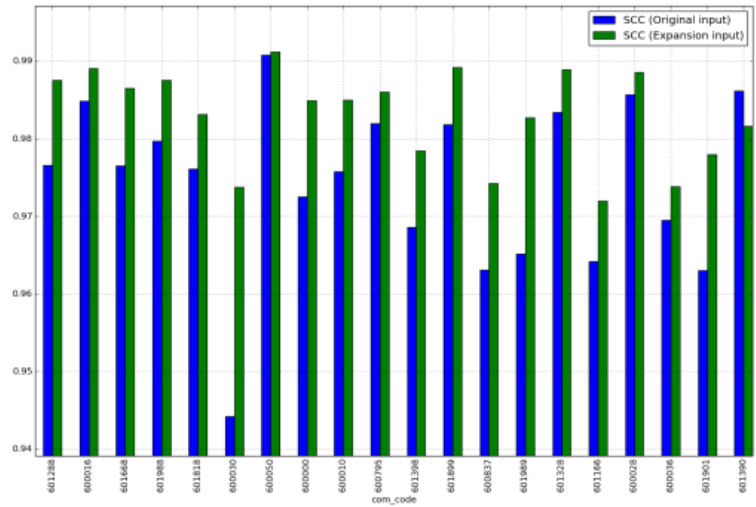

Fig. 6. SVR result for different input data type.

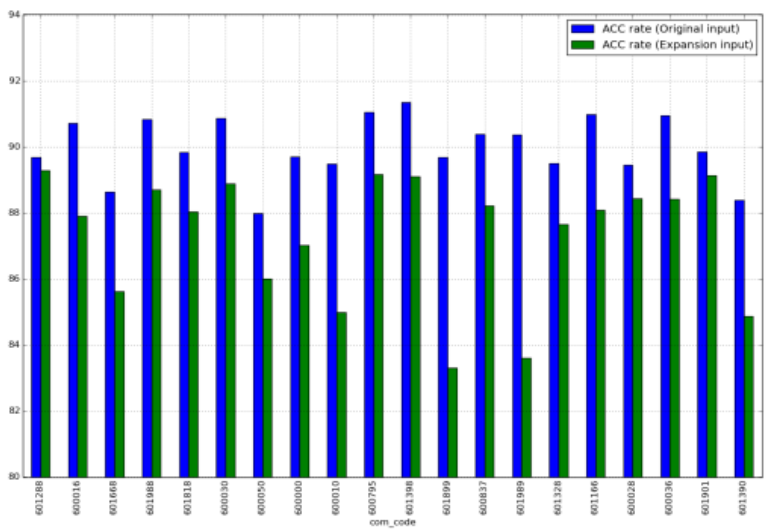

Fig. 7. SVC result for different input data type.

With no accident, expansion input model preforms better than original model in general except that the company *601390* in SVR, whereas in SVC model the result is entirely opposite. One possible reason for that is the sentiment dictionary is partial. That means 0 is a not neutral score if the scoring group score words all in a positive way or negative way. As in this experiment we set all news input nodes to be 0 to represent that the day has no news, 0 could stand for a negative or positive result in the model finally. Such kind of data can be an assignable disturbance in input data, which will obviously result a partial prediction. This is especially obvious in classification as there is only one hyperplane. Such a small disturbance can also result a large error in the end.

### 3.1.3. General SVM result

SVR indeed has a good fitting ability when predicting stock price. Figure below shows the SVR fitting result for *bank of China (601988).* The model's SCC is 98.5429% and its MSE is 0.00328.

It can be seen from the figure that SVR is excellent in forecasting the sharp changes, which is hard for most linear algorithms. However, it may also have one day or two's delay. That means when real stock market's price is already descending, the predicting result is just about to change its trend. That maybe does something with the not good enough SVC result. The model needs to use some real stock data to make sure the tendency of curves.

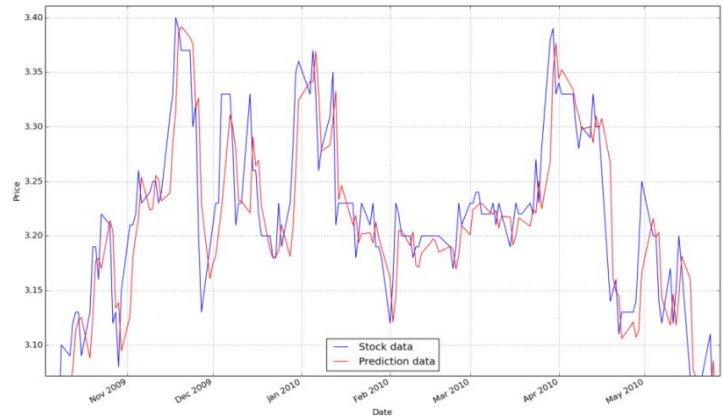

Fig. 8. SVR result of 601988 (10/1/2009-6/1/2010).

As for SVC result of 601988, the accuracy is 59.1734%. The result fairly cannot say be good. It is also one latent obstacle for more elaborate SVR. However, as discussed above, stock market is unable to predict entirely. Compared to some previous work, which is shown in the below figure, our research has made some progress.

Table 2. Some stock Price Forecasting Research Results

| *Research* | *Whether use news* | *MSE (regression)* | *ACC (classification)* |
|---|---|---|---|
| *This research* | *Yes* | *0.00328* | *59.17%* |
| *Tang et al, 2009[13]* | *Yes* | \ | *0.51% & 50%* |
| *Mittermayer 2004[14]* | *Yes* | \ | *58%* |
| *Zhai et al, 2007[15]* | *No* | \ | *58.80%* |

In general, the majority of researches are interested in predicting the trends of stocks instead of price forecasting. One significant reason is that the regression result is quite good enough (SCC reaches 99% degree). There is not too much academic value to raise the index to a little higher. For the classification aspect, main research stream's results (ACC) are around 60%. Most of them stand at the interval from 50% to 60%. Some reach higher even to 70% (like [16]). However, their dataset are relatively partial, in limited amount, categories and specific time, etc.

### 3.1.4. Parameter optimization

In LIBSVM, two universally accepted parameters: c and g need to be set at first, which represent the cost and gamma in kernel function respectively. Based on the fact that the default search tool's performance is not good, this research designs two searching methods based on the core idea of grid search method and genetic algorithm. After testing, this research set search domains for c and g are [1, 25] and [0.01, 0.30] respectively. Their search step lengths are set to be 1 and 0.01.

### 3.1.5. Traverse algorithm

If traversing the whole solution space, there is always a best parameter pair to perform in the model. The different part from the LIBSVM's grid search is the evaluation process. For each *(c, g)* pair, 10 random train dataset and test dataset groups are picked from database. SCC and MSE value of 10 groups will be averaged for evaluating. In such situation, the average SCC and MSE have no statistic meaning and only serve to the evaluation purpose. The optimum solution plane of traverse result is shown below.

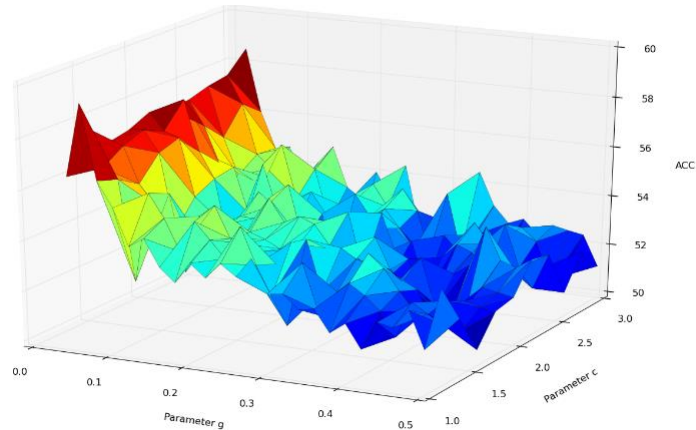

Fig. 9. Traverse algorithm best parameter plane (SVC).

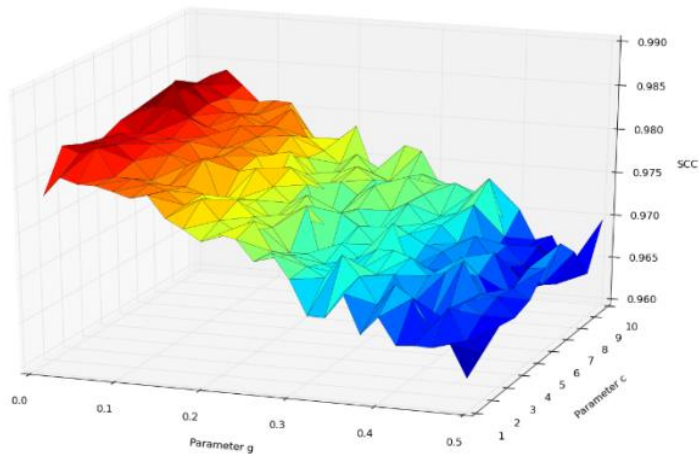

Fig. 10. Traverse algorithm best parameter plane (SVR).

The figures also show that *g* is the main factor in the model which proves that searching *g* first is correct in approximate algorithm discussed the latter. Actually, when *g* is near 0, the model will perform much better. For all 20 stocks, the traverse algorithm result is shown below.

Table 3. Traverse Algorithm's result of All Stocks for SVC

| Com_code | Best g | Best c | Best ACC (%) | Com_code | Best g | Best c | Best ACC(%) |
|---|---|---|---|---|---|---|---|
| 601288 | 0.02 | 1.4 | 60.16462 | 601398 | 0.02 | 1.6 | 54.83435 |
| 600016 | 0.3 | 1.6 | 53.10473 | 601899 | 0.28 | 2.4 | 54.74664 |
| 601668 | 0.48 | 2 | 53.65812 | 600837 | 0.16 | 2.8 | 51.43552 |
| 601988 | 0.04 | 1.2 | 59.1734 | 601989 | 0.06 | 2.2 | 55.49496 |
| 601818 | 0.02 | 1.4 | 59.12176 | 601328 | 0.02 | 1.6 | 57.57212 |
| 600030 | 0.44 | 1.2 | 53.36816 | 601166 | 0.08 | 1.2 | 53.47609 |
| 600050 | 0.26 | 2.6 | 53.95256 | 600028 | 0.02 | 2.4 | 52.82336 |
| 600000 | 0.28 | 2.4 | 51.70183 | 600036 | 0.02 | 1.2 | 52.86296 |
| 600010 | 0.16 | 2.4 | 53.23738 | 601901 | 0.34 | 1.8 | 52.84631 |
| 600795 | 0.02 | 3 | 56.99535 | 601390 | 0.02 | 1.8 | 59.3652 |

Table 4. Traverse Algorithm's Result of All Stocks for SVR

| Com_code | Best g | Best c | Best MSE | Best SCC | Com_code | Best g | Best c | Best MSE | Best SCC |
|---|---|---|---|---|---|---|---|---|---|
| 601288 | 0.02 | 4 | 0.002898 | 0.982576 | 601398 | 0.02 | 7 | 0.003794 | 0.975532 |
| 600016 | 0.02 | 9 | 0.02135 | 0.992012 | 601899 | 0.04 | 10 | 0.012369 | 0.990894 |
| 601668 | 0.06 | 4 | 0.009668 | 0.98962 | 600837 | 0.06 | 10 | 0.143558 | 0.987999 |
| 601988 | 0.04 | 7 | 0.00328 | 0.985429 | 601989 | 0.02 | 9 | 0.027772 | 0.986388 |
| 601818 | 0.02 | 10 | 0.006065 | 0.984559 | 601328 | 0.04 | 5 | 0.01065 | 0.990556 |
| 600030 | 0.04 | 9 | 0.226013 | 0.991033 | 601166 | 0.04 | 8 | 0.064489 | 0.987322 |
| 600050 | 0.08 | 8 | 0.009361 | 0.99278 | 600028 | 0.04 | 10 | 0.020217 | 0.992863 |
| 600000 | 0.04 | 6 | 0.061748 | 0.991011 | 600036 | 0.04 | 9 | 0.053716 | 0.981747 |
| 600010 | 0.04 | 8 | 0.009559 | 0.98801 | 601901 | 0.04 | 7 | 0.074697 | 0.989365 |
| 600795 | 0.02 | 9 | 0.003774 | 0.985014 | 601390 | 0.04 | 8 | 0.023091 | 0.993408 |

As we can see from the tables, best parameter set *(c, g)* of 20 stocks for SVC are differential, while for SVR

they are almost all in interval (0.2,0.4). In comparison, best *g* fluctuates a lot when stock changes in SVC, which means that investors have quite different sensitivities for different companies. Even if there is same news' impact on different companies, the result will be different. That might do something with investors' trust and confidence for different companies.

### 3.1.6. Approximate algorithm

This algorithm will divide random 50 train dataset and test dataset groups. For each group, traverse algorithm will be used to find its own best pair. Surely for different groups the experiment result will be different. Each best pair is one of the local optimums. The thought is that when the group number is getting larger, there surely is a trend that these local optimums will approximate global optimum. It can be deemed that approximate results are near each other, surrounding our global optimum.

As assumed, experiment result shows that the approximate algorithm final result for all stocks is also similar with traverse algorithm. Fig. 11 shows the local optimums of approximate algorithm and the global optimum of traverse algorithm in SVC model. With step length becoming smaller, the convergence precision will increase. Thus, the evaluation process of traverse algorithm is reasonable. Considering the time complexity, traverse algorithm is a better option.

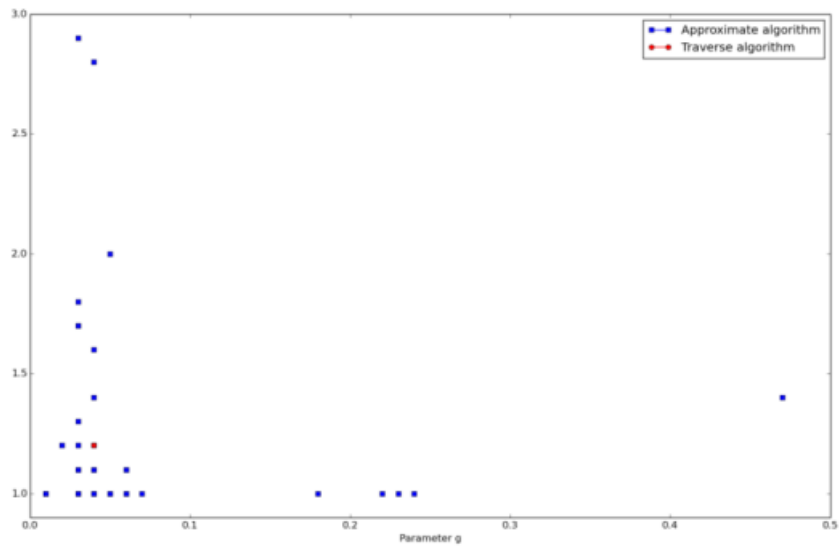

Fig. 11. Two algorithms' results for SVC.

## 4. Source Impact Factors

As there are X input nodes representing the news source, it is possible to explore the influence of each source by measuring the predicting results' difference from the normal prediction result, when giving the same fluctuation in the input node. Theoretically speaking, when the same deviation is given, the source whose result has bigger gap to the normal prediction has a higher weight.

For each stock data, the fifth piece of data will be chosen as the test data. Each time 100 will be added to the every source node's value to predict. By comparing the normal prediction, source weight for each stock can be obtained. The calculate formula is shown below.

$$M_{ij} = {p_{ij} - p_{i0}}/{p_{i0}}$$

where $M_{ij}$ represents *No.j* sources' weight for *No.i* stock, $p_{ij}$ represents prediction price with *No.j* news nodes' value adding 100 and $p_{i0}$ represents the normal prediction price.

Considering that each stock has different importance in stock market, the source weight for whole stock market using the 20 stocks result is shown below.

$$Z_j = \sum_{i=1}^{20} M_{ij} \cdot V_i$$

where $V_i$ represents the trading volume of *No.i* stock.

After scaling, the final source weight is shown in Table 5.

Table 5. Source Weight

| No. | Source | Weight | No. | Source | Weight |
|---|---|---|---|---|---|
| 1 | Wallstreetcn | 19.4154 | 11 | ccstock | 18.78984 |
| 2 | Stockstar | 19.37817 | 12 | Cnstock-SH | 18.78599 |
| 3 | Chinanews | 19.08397 | 13 | Cnstock | 18.73956 |
| 4 | Cnstock & CS | 19.03179 | 14 | Cnstocknet | 18.73758 |
| 5 | 21st CBH | 19.01998 | 15 | STCNnet | 18.73686 |
| 6 | NBD | 18.95937 | 16 | Eastmoney | 18.71921 |
| 7 | Yicai | 18.95242 | 17 | Xinhuanet | 18.7151 |
| 8 | Tencent Finance | 18.90835 | 18 | NNBD | 18.65674 |
| 9 | Xinlang Finance | 18.88499 | 19 | Aastocks | 18.65194 |
| 10 | STCN | 18.87678 | 20 | Hexun | 18.64479 |

The result is reasonable relating to Chinese stock market case. It seems there are two major factors determining the source's weight. One is the news quality. Like the *wallstreetcn, chinanews* and 21st *CENTURY BUSINESS HERALD*, they are all famous for its high news quality. The other is the audience number. Actually *Tencent Finance* and *Xinlang Finance* are not the original birthplace of news documents. Just because they have huge amount of audience, they can stand so front in this rating. In addition, the main high weight sources come from the newspapers and magazines, showing that the majority of Chinese investors are still depend on paper news as their financial information source.

## 5. Conclusions and Discussions

In general, the SVM method based on text mining technology shows an outstanding result in predicting stock market especially when predicting the specific stock price. In comparison, when predicting the stock tendency, SVC is not good enough, though it also proves the importance of news documents.

In addition, this paper proposes two simple and basic parameter-find algorithms and discusses the relationship between them. The result shows that traverse algorithm is better. Furthermore, based on the test best parameter-searching plane, in SVM parameter g is the main factor and one can just choose search g first in order to increase searching speed.

This paper also proposes a simple approach to evaluate the online sources' weight in affecting investors' trading decision-making. The weight result is based on the real effect and is surely much better than methods based on the audience number or specialists' proposal.

However, the SVM is not perfect yet. When news amount is low, the predicting result is not good enough. In such cases, even some basic and simple linear regression algorithm can do a better job.

There are also several aspects needing to make sure and improve in order to get a more robust SVM method. 1) Adding more text sources; 2) Designing a standard sentiment evaluation system and finding more specialists to score the sentiment dictionary; 3) Making sure the stock needing to be predicted has enough text documents and trading volume; 4) Trying to expand the whole dataset.

## References

[1] Bollerslev, T. (1986). Generalized autoregressive conditional heteroskedasticity. *Journal of Econometrics, 31*, 307–327.
[2] Peiris, S., Allen, D., & Yang, W. L. (2005). Some statistical models for durations and an application to news corporation stock prices. *Mathematics and Computers in Snimulation, 68*, 549-556.

[3] Wang, Y. H. (2009). Nonlinear neural network forecasting model for stock index option price: Hybrid GJR-GARCH approach. *Expert Systems With Applications, 36*, 564-570.

[4] Chang, J. F., Wei, L. Y., & Cheng, C. H. (2009). Anfis-based adaptive expectation model for forecasting stock index. *International Journal of Innovative Computing, Information and Control, 5(7),* 1949-1958.

[5] Geva, T., & Zahavi, J. (2014). Empirical evaluation of an automated intraday stock recommendation system incorporating both market data and textual news. *Decision Support Systems, 57*, 212-223.

[6] Nizer, P. S. M., & Nievola, J. C. (2012). Predicting published news effect in the Brazilian stock market. *Expert Systems With Applications, 39,* 10674-10680.

[7] Peiris, S, Allen, D, & Yang, W. L. (2005). Some statistical models for durations and an application to News Corporation stock prices. *Mathematics and Computers in Snimulation, 68*, 549-556.

[8] Ghiassi, M., Skinner, J., & Zimbra, D. (2013). Twitter brand sentiment analysis: A hybrid system using n-gram analysis and dynamic artificial neural network. *Expert Systems With Applications, 40*, 6266–6282.

[9] Luss, R., & D'Aspremont, A. (2015). Predicting abnormal returns from news using text classification. *Quantative Finance, 15*, 999-1012.

[10] Wu, D. D., Zheng, L. J., Olson, D. L. (2014). A decision support approach for online stock forum sentiment analysis. *IEEE Transactions on Systems Man Cybernetics-Systems, 44*, 1077-1087.

[11] Schumaker, R. P., Zhang, Y., Huang, C. N., & Chen, H. (2012). Evaluating sentiment in financial news articles. *Decision Support Systems, 53*, 458-464.

[12] Abdullah, S. S., Rahaman, M. S., & Rahaman, M. S. (2013). Analysis of stock market using text mining and natural language processing. *Proceedings of International Conference on Informatics, Electronics and Vision (ICIEV)*, Univ Dhaka, Dhaka, Bangladesh.

[13] Tang, X., Yang, C., & Zhou, J. (2009). Stock price forecasting by combining news mining and time series analysis. *Proceedings of 2009 IEEE/WIC/ACM International Conference on Web Intelligence and Intelligent Agent Technology Workshops*.

[14] Mittermayer, M. A. (2004). Forecasting intraday stock price trends with text mining techniques. *Proceedings of the 37th Hawaii International Conference on System Sciences*.

[15] Zhai, Y., Hsu, A., & Halgamuge, S. K. (2007). Combining news and technical indicators in daily stock price trends prediction. *Advances in Neural Networks, 4493*, 1087-1096.

[16] Kao, L. J., Chiu, C. C., Lu, C. J., & Yang, J. L. (2013). Integration of nonlinear independent component analysis and support vector regression for stock price forecasting. *Neurocomputing, 99*, 534-542.

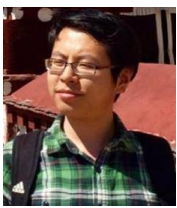
**Yancong Xie** is a current management information systems PhD student at Information Systems School, Queensland University of Technology. He received his bachelor degree in mathemaics from Renmin University of China in 2015. This paper is formed based on his undergraduate graduation thesis. Yancong's research interests include data mining, mathematical modeling, simulation, methodology, theory building, and information systems impact.

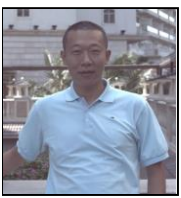
**Hongxun Jiang** is an associate professor at School of Information, Renmin University of China. He received a B.S. degree and M.S. degree in information system engineering, Ph.D. degree in management science and engineering, all from School of Economics and Management, Beihang University. His research interests include data science, finance engineering, and optimize.